\documentclass{article}
\PassOptionsToPackage{table}{xcolor}
\usepackage{iclr2027_conference,times}

\usepackage{latexsym}
\usepackage[T1]{fontenc}
\usepackage[utf8]{inputenc}
\usepackage{microtype}
\usepackage{amsmath,amsfonts,bm}

\def\eqref#1{equation~\ref{#1}}
\def\1{\bm{1}}

\DeclareMathAlphabet{\mathsfit}{\encodingdefault}{\sfdefault}{m}{sl}
\SetMathAlphabet{\mathsfit}{bold}{\encodingdefault}{\sfdefault}{bx}{n}

\usepackage{xcolor}
\definecolor{hzBuilder}{HTML}{2A78D6}
\definecolor{hzInk}{HTML}{0B0B0B}
\definecolor{hzViolet}{HTML}{4A3AA7}
\usepackage[colorlinks=true,citecolor=hzBuilder,linkcolor=hzInk,urlcolor=hzViolet]{hyperref}
\usepackage{url}
\usepackage{xurl}
\usepackage{booktabs}
\usepackage{graphicx}
\usepackage{pifont}

\title{Humanize Olympiad Agents: A Multi-Session Workflow for Olympiad Problem Solving with Solve, Review, Wiki, Verify}
\author{\textbf{Zhengyang Zhang\textsuperscript{1},
Jing Xiong\textsuperscript{3},
Hongzhou Lin\textsuperscript{6},
Juihui Chung\textsuperscript{5}} \\
\textbf{Zijian Zhang\textsuperscript{1,2},
Song Han\textsuperscript{2,4},
Sihao Liu\textsuperscript{2},
Ligeng Zhu\textsuperscript{2}} \\[0.5em]
{\small \textsuperscript{1} Tsinghua University \quad
\textsuperscript{2} NVIDIA \quad
\textsuperscript{3} The University of Hong Kong} \\
{\small \textsuperscript{4} MIT \quad
\textsuperscript{5} Princeton University \quad
\textsuperscript{6} Amazon }
}

\newif\ifshowauthors
\showauthorstrue
\ifshowauthors\iclrfinalcopy\fi
\begin{document}

\maketitle
% \lhead{Under review as a conference paper at ICLR 2027}
\lhead{Preprint}

\begin{abstract}
Large language model (LLM) agents are rapidly improving on long-horizon scientific and olympiad tasks. Recent systems have produced a Lean-formalized resolution of the Navier--Stokes problem and autonomous solutions to four open questions in the Erd\H{o}s Conjectures database \citep{openai2026navierstokes,google2026aletheia}. Yet neither system releases both its underlying model and complete inference stack. More broadly, a single agent session remains brittle on difficult problems: an early error can propagate, missing domain knowledge can stall progress, and the agent can appear to succeed after misreading or weakening the task. We introduce \textbf{HOA (Humanize Olympiads Agent)}, an open, flow-level framework that coordinates multiple agent sessions through three complementary feedback channels. First, an independent reviewer checks each candidate and returns actionable feedback. Second, a structured wiki supplies task-specific knowledge that may be missing or unreliable in model pretraining. Third, a verifier  checks generated Lean proofs against trusted challenge statements.
% \ligeng{HOA flow-level, mutli-session, extendable agents, line 21-22}.
% Together, these components turn one-shot generation into an iterative cycle of generation, critique, knowledge retrieval, revision, and verification. 
With GPT-5.6 Sol, HOA reaches the \textbf{perfect score on  IMO, IPhO, and IOI 2026}, and \textbf{near-full score on IChO 2026} problem set and solves \textbf{all 672 problems (100\%) on PutnamBench}. On the Lean-Eval leaderboard, HOA  ranks \textbf{\#1 with 251 of 304 problems solved (82.6\%)} \citep{leaneval}. Furthremore, we extend HOA's results with the open models -- With Kimi-K3, HOA also achieves full score on IMO, IPhO,on IOI, and gold-level 97.1\% on IChO, which closely matches the closed-source GPT-5.6 Sol + \texttt{/Goal} baseline on IMO and IPhO and exceeds it on IChO and IOI, showing that systematic feedback can compensate for substantial differences in base-model capability. %\ligeng{numbers}
\par\noindent\textbf{Code:} \url{https://github.com/humanfia/hoa}

\end{abstract}

\section{Introduction}

LLM-based agents can now carry out long sequences of reasoning and action. They browse and synthesize evidence, develop and test code, construct mathematical arguments, and interact with formal proof assistants. Systems such as Deep Research, the AI Scientist, and AI co-scientist show how tools and iterative orchestration can turn a language model into a research workflow rather than a single-response generator \citep{openai_deep_research,lu2024aiscientist,google2025coscientist}. Their performance depends on both the base model and the \emph{harness}: the prompts, tools, context management, execution environment, and control flow that allocate the inference budget, but all these works do not reveal those details. Community thereby needs open harnesses for those reproducible results.

Yet most evaluations treat a single agent session as the basic unit of problem solving. This design can be easily integrated into the next generation model and  is fragile for hard research-level problems.  % \zhengyang{}
% 不能放大模型能力，被post train 吃进去
First, a plausible but false intermediate claim can contaminate the rest of a long trajectory. Second, more reasoning cannot supply a theorem, scientific rule, or algorithmic pattern that the model does not know. Third, an agent may misread a requirement or weaken a goal---for example, by proving a nearby statement or testing convenient cases---and appear to succeed without solving the original task. We refer to this last failure mode as problem misunderstanding and reward hacking. These failures make longer single trajectories yield sharply diminishing returns.

We distinguish three levels of agents. At the \emph{model level}, a direct API call returns an answer without an execution environment. At the \emph{tool level}, a CLI-harness such as Codex and Claude Code gives the model tools to read files, execute code, and check intermediate results. At the \emph{flow level}, a workflow coordinates multiple agent sessions and tools. Humanize is a flow-level protocol that extends Codex with an independent reviewer and a worker--reviewer revision loop \citep{humanize_rlcr}. The reviewer helps detect hallucinations, problem misunderstanding, and reward hacking.

\begin{figure}[t]
  \centering
  \includegraphics[width=0.9\linewidth]{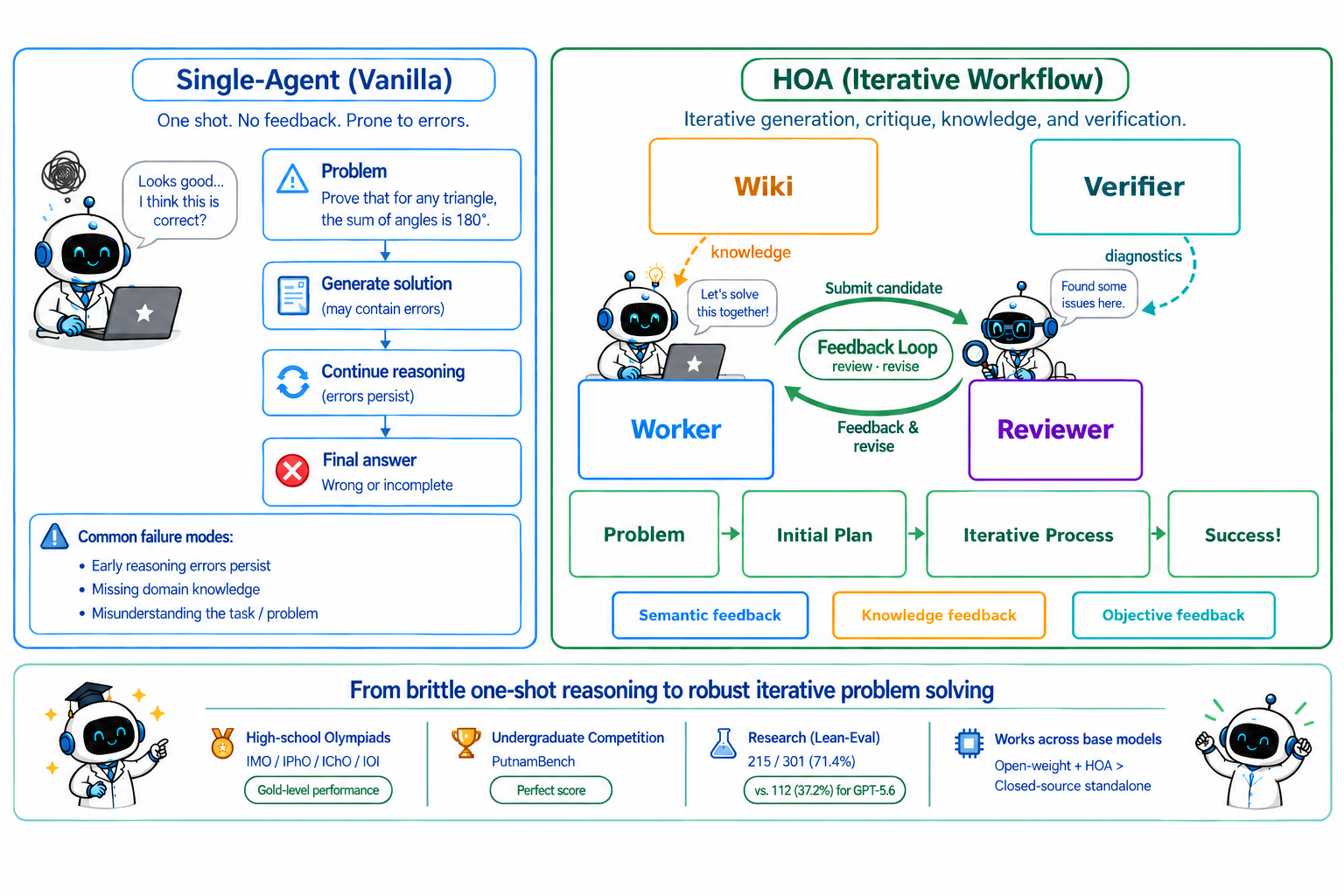}
  \vspace{-15pt}
  \caption{HOA replaces a brittle one-shot workflow with iterative review, knowledge retrieval, and verification. The bottom row summarizes results from high-school olympiads to research-level formal mathematics. 
  %\ligeng{Use codex to make an editable version in keynote or powerpoint.} \ligeng{why HOA is better previous harness?
  %1. multi-sessions, long horizon running
  %2. worker-reviewer mechanism, avoid halluciation
  %3. extentable for wiki and verifications, introduce new domain knowledge and ensure correct solutions \zhengyang{Problem, no captical}
  %}
  }
  \vspace{-12pt}  
  \label{fig:teaser}
\end{figure}

\begin{figure}[t]
  \centering
  \includegraphics[width=0.9\linewidth]{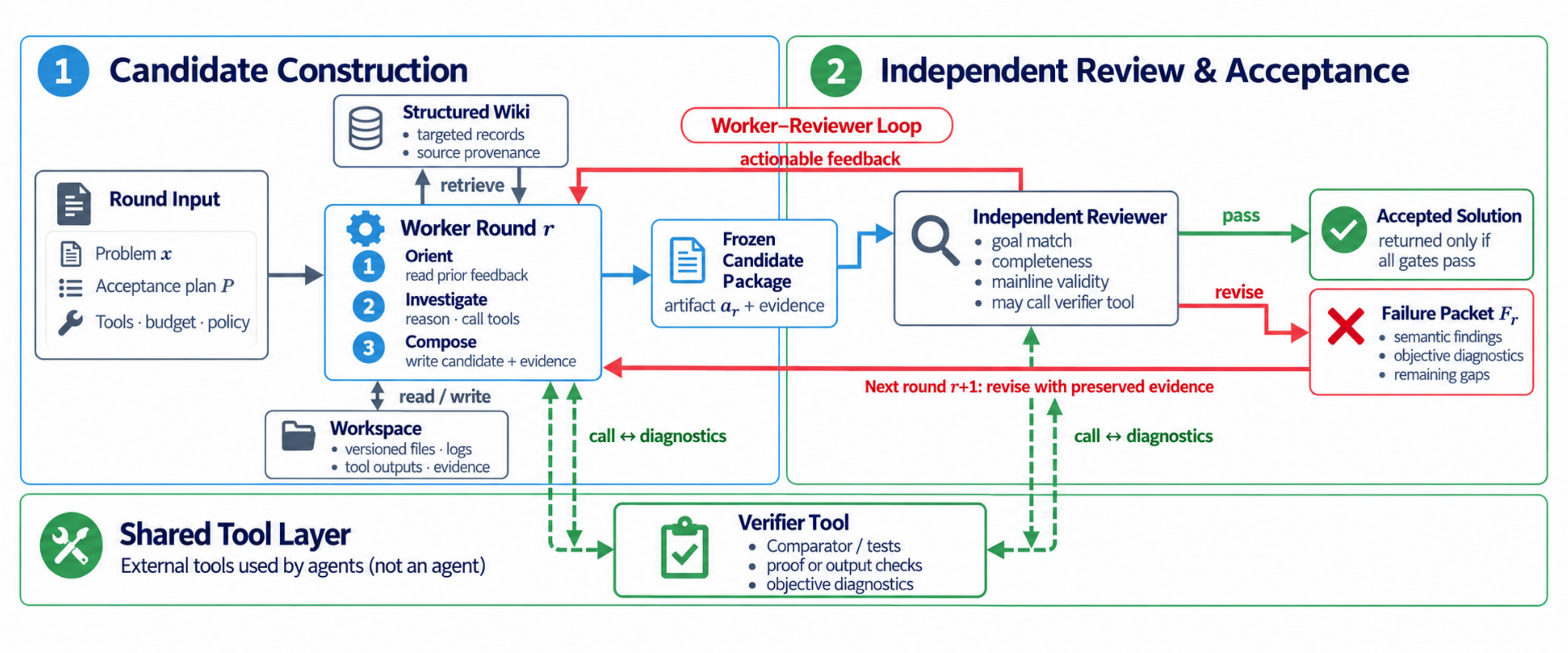}
  \vspace{-18pt}
  \caption{Overview of HOA. The worker receives the problem, acceptance plan, tools, budget, and prior feedback, and may retrieve provenance-bearing wiki records while building a versioned candidate. An independent reviewer inspects the frozen candidate and returns semantic findings. The worker and reviewer may both call the external verifier for objective checks. If any requirement fails, the findings and diagnostics start the next bounded worker round.}
  \vspace{-10pt}
  \label{fig:overview}
\end{figure}

We propose \textbf{HOA (Humanize Olympiads Agent)}, whose new contribution is an artifact-centered acceptance protocol that routes structured knowledge retrieval, independent semantic review, and task-specific objective verification through a shared worker--reviewer repair loop and accepts a candidate only after all applicable gates pass. A worker constructs a candidate artifact and consults a structured wiki when it identifies a knowledge gap. An independent reviewer checks the frozen artifact against the original problem and acceptance plan, then returns specific findings to the worker. Semantic review and objective validation remain separate gates. For mathematical tasks, the reviewer runs verifier on the unchanged candidate before accepting it \citep{lean_comparator}. A result is accepted when every required gate passes. Figure~\ref{fig:teaser} and Figure.~\ref{fig:overview} explains how HOA solves a complex olympiad question.

The three feedback channels address different failure modes. Review catches omitted cases, invalid inferences, and solutions to the wrong problem. The wiki supplies missing knowledge. Verification provides an objective, task-specific signal. HOA keeps these roles separate while applying all feedback to the same evolving artifact.
We evaluate HOA on high-school olympiads, college-level competition mathematics, and research-level formal mathematics, and HOA achieves top performance in various competitions and benchmarks. On the Lean-Eval leaderboard, HOA ranks \#1 with 251/304 solves (82.6\%) \citep{leaneval}.
% HOA achieves perfect score and full mark on several benchmarks and ranks top in lean-eval with GPT-5.6-sol.  
% Most notably, we also evaluate HOA with several open source models and achieve close performance with commercial ones. This result shows that workflow design can boost the models' performance and close the gap between open- and close- models.
% Divide the contributions into sentences.

Our contributions are:

\begin{itemize}
  \item We introduce a flow-level worker--review framework for olympiads and frontier problems solving.  We add task-specific  verification tools, together with a structured wiki that supplies targeted, provenance-tracked knowledge, and thoroughly evaluate across various domains.
  % Wiki+ verfifier + workflow 本身，分别解决了 3个问题
  % No multiagent
  \item We introduce a flow-level formalization agents that can turn physics problems and figures into typed Lean statements, checks them against the original problems, to enable later agent-driven proof. %\ligeng{@xiongjing, draft some details about the formalization}
  \item HOA reaches state-of-the-art results: \textbf{full scores on IMO, IPhO, IOI 2026}, and \textbf{near pefect score on IChO 2026}, all based on the latest competitions and there is no leakage. HOA also solved \textbf{672/672 on PutnamBench} including all problems from 1963 to 2023, and achieves the \textbf{\#1 ranking on the Lean-Eval leaderboard}.
  \item  HOA also shows the potential of open harness plus open models. We extensively test open source models and shows that HOA can boost the performance of open models and enables open-weight models to match or surpass standalone closed-source agents.

  % 1. flow
  % 2. besides flow
  % 3. open model
  
\end{itemize}

\section{Related Work}
\label{sec:related-work}

\subsection{Language Models for Formal Theorem Proving}

LLM-based theorem proving in Lean \citep{moura2021lean} includes specialized provers, tool-using agents, and multi-stage systems. Specialized provers include TheoremLLaMA, Hunyuan-Prover, the DeepSeek-Prover series, the Goedel-Prover series, and Kimina-Prover \citep{wang2024theoremllama,li2024hunyuanprover,xin2024deepseekprover,xin2024deepseekv15, ren2025deepseek,lin2025goedel,lin2025goedelv2,wang2025kimina}. Lean-STaR interleaves reasoning and proof tokens during generation \citep{lin2024lean}, while AlphaProof uses RL-based proof search at olympiad scale \citep{alphaproof}.

Tool-using agents query Lean during inference and revise failed proofs. AxProverBase and Numina-Lean-Agent use compact agent loops \citep{requena2026minimal,liu2026numina}, while LongCat-Flash-Prover integrates tool use into reinforcement learning \citep{wang2026longcat}. OProver repairs failed proofs with compiler feedback and retrieved examples \citep{ma2026oprover}, and VERITAS routes verifier feedback into critic-guided search \citep{acharya2026veritas}. Other systems add explicit proof plans: Draft-Sketch-Prove translates informal proofs into formal sketches \citep{jiang2022draft}; Hilbert recursively decomposes goals \citep{hilbert2025}; Seed-Prover refines proof sketches \citep{seedprover2025,seedprover15}; Goedel-Architect revises a global dependency graph \citep{chung2026goedelarchitectstreamliningformaltheorem}; and LEAP combines informal blueprints, decomposition, self-refinement, and Lean feedback \citep{kung2026leap}. HOA adds an independent worker--reviewer loop, reusable knowledge, and task-specific verification across formal mathematics and multidisciplinary olympiad tasks.

\subsection{Long-Horizon Research and Olympiad Agents}

Research agents extend language models with browsing, code execution, and iterative control. Deep Research supports extended browsing and synthesis \citep{openai_deep_research}; the AI Scientist covers ideation, experimentation, writing, and simulated review \citep{lu2024aiscientist}; and AI co-scientist uses specialized generation, reflection, and ranking stages to refine hypotheses \citep{google2025coscientist}. AlphaEvolve pairs LLM-generated programs with evolutionary search and automatic evaluators \citep{novikov2025alphaevolve}. MLE-bench further shows that both the base model and agent scaffold shape long-horizon engineering performance \citep{chan2024mlebench}.
% Add google 2025 IOI

Olympiad and research agents increasingly combine generation with iterative verification. Aletheia uses a generator--verifier--reviser loop, browsing, and inference-time scaling \citep{google2026aletheia}. A model-agnostic verification-and-refinement pipeline reaches IMO gold-level performance \citep{li2025imogold}, while domain-specific scaffolds support chemistry and broader science olympiads \citep{chemlabs2025}. RMA combines literature search, a shared knowledge bank, and multi-round proposer--verifier feedback for research mathematics \citep{zhao2026rma}. In programming, Gemini 2.5 Deep Think solved 10 of 12 problems at the 2025 ICPC World Finals, a gold-medal-level result that would have ranked second among the competing teams \citep{google2025icpc}. Generation, behavioral clustering, and ranking have also reached gold-level IOI 2025 performance with open-weight models \citep{ioi2025openweight}. HOA combines an independent worker--reviewer repair loop, a structured wiki, and task-specific verification in one workflow, then applies it across mathematics, physics, chemistry, and programming.

\subsection{Prover Driven by Agent Loops}

Formal provers increasingly use long-running agent loops. AxiomProver generated Lean proofs for all 12 Putnam 2025 problems, although its public release provides proof artifacts rather than a full system description \citep{axiom2026prover}. Aleph Prover plans subproblems, checks them in Lean, and updates a global plan; paired with GPT-5.2, it reports solving 668 of 672 PutnamBench problems \citep{logicalintelligence2026aleph}. These systems show that repeated planning and verification can scale formal reasoning.

HOA builds on Humanize~\citep{humanize_rlcr}, in which an independent worker and reviewer repeat until explicit acceptance criteria are met. Related methods use self-critique, stored feedback, or tool-grounded critique~\citep{madaan2023selfrefine,shinn2023reflexion,gou2023critic}, but self-correction can fail without a reliable external signal \citep{huang2024selfcorrect}. HOA therefore separates generation, review, and verification: the reviewer provides evidence and specific revision requests, and the worker resubmits the revised artifact. For Lean tasks, Comparator builds the submission in a sandbox, matches it against the trusted statement, restricts permitted axioms, and checks the proof with the Lean kernel \citep{lean_comparator}. These checks matter because a proof can compile while using a disallowed axiom \citep{vamshi2026rewardoracle}.

% merge 2.4 with 2.2

Agent loops also benefit from external knowledge. Retrieval-augmented generation, ReAct, and RARR combine model reasoning with retrieved evidence and external actions \citep{lewis2020rag,yao2023react,gao2023rarr}; LeanDojo and ReProver retrieve premises from Lean libraries \citep{yang2023leandojo}. HOA instead uses a bounded, task-specific wiki. Following Tau Ceti, Lean Pool, and DreamProver \citep{tauceti,leanpool,zhang2026dreamprover}, its mathematics wiki turns accepted proofs into provenance-tracked lemma records. Its chemistry wiki draws from textbooks, olympiad problems outside the evaluation set, and online references. Each run retrieves a small set of relevant records.

\section{Method}
\label{sec:method}

\subsection{Overview}

HOA is a flow-level agent framework built on Humanize \citep{humanize_rlcr}. It contains a \emph{worker}, an independent \emph{reviewer}, a structured \emph{wiki}, and a task-specific \emph{verifier}. Let $x$ denote the original problem and $a_t$ the artifact produced at iteration $t$. The worker constructs $a_t$ from $x$ and prior feedback, consulting the wiki when it can state a specific knowledge need. The reviewer checks semantic correctness, while the verifier checks a machine-evaluable condition. For mathematics and physics tasks, HOA uses Comparator as the verifier \citep{lean_comparator}. A candidate counts as solved after every required gate accepts it.

HOA combines three forms of feedback. Semantic feedback asks whether the artifact fully answers the original problem. Knowledge feedback supplies missing facts or methods. Objective feedback tests a machine-evaluable condition. These signals are complementary rather than interchangeable.

%\paragraph{Budgets and termination.}
%For every experiment we fix the maximum number of cycles, total model tokens,

%tool calls, and wall-clock timeout before starting the run.  A run that timesout, exceeds its budget, or never obtains both acceptance signals is counted as unsolved.

\subsection{Worker--Reviewer Loop}
%\ligeng{worker-reviewer loop into a single paragraph}

% \paragraph{Overview.}
The worker--reviewer loop separates artifact construction from acceptance. In each round, the worker produces or revises a candidate. An independent reviewer then evaluates the frozen candidate against the original problem and acceptance criteria. Unresolved findings return to the worker as actionable feedback for the next round. The loop ends when the reviewer declares the candidate complete or the run reaches its resource limits. This separation lets the worker explore and repair without judging its own output.

The worker maintains the evolving artifact. In the first round, it receives the problem, task instructions, available tools, and acceptance criteria. In later rounds, it also receives the current artifact, reviewer findings, and verifier diagnostics. The worker must address every reported issue. When it identifies a knowledge gap, it sends a focused wiki query, incorporates the retrieved material with provenance, and revises the artifact. It may call the verifier for immediate machine feedback, but it cannot accept its own output or bypass review.
% to distinguish retrieval failures
% from reasoning failures in the analysis.

The reviewer receives the original problem, candidate artifact, acceptance plan, and a runner-generated summary of the current round. It does not see the worker's hidden reasoning or conversation history, and the candidate remains read-only during review. This separation reduces anchoring and prevents silent repairs. The reviewer instead returns findings that guide the next worker round.

% The HOA protocol classifies findings as \emph{Mainline Gaps}, \emph{Blocking Side Issues}, or \emph{Queued Side Issues}. It also assigns a \emph{Mainline Progress Verdict} of \textsc{advanced}, \textsc{stalled}, or \textsc{regressed}. The reviewer ends its report with \texttt{COMPLETE} when the original plan and all acceptance criteria are satisfied. Otherwise, its findings start the next worker round.

\subsection{Wikis for Domain-Specific Knowledge}

HOA uses separate wikis for mathematics and chemistry because the two domains require different forms of reusable knowledge.

\paragraph{Mathematics wiki.}
The mathematics wiki follows the reusable formal libraries developed by Tau Ceti and Lean Pool \citep{tauceti,leanpool}. After a mathematical problem is solved and accepted, a curation agent extracts the main theorem and reusable intermediate lemmas from the Lean development. Each record contains declaration names, Lean signatures, a short description, the source problem, required imports, source locations, and verification status. 

\paragraph{Chemistry wiki.}
The chemistry wiki is built before evaluation from textbooks, olympiad problems outside the test set, and relevant online references. Agents organize the material into concept-centered pages that cover definitions, reactions, formulas, experimental procedures, problem types, and solution patterns. Each page records its sources.
%For copyrighted or access-restricted material,
%we store only basic source information and our own summary, not the original text.
%The held-out test questions, solutions, evaluation transcripts, and expert
%reasoning paths are excluded from the corpus.
%\ligeng{why no physic / bio / ioi wiki}

We do not build separate wikis for physics or coding. IPhO and IOI required knowledges are largely already presented in model pretraining. We focus the wiki on mathematics and chemistry, where structured external knowledge provides greater value.

When the worker identifies a missing fact or method, it sends a focused query. The retrieval layer searches Lean declarations in the mathematics wiki or concept pages in the chemistry wiki, then returns a small set of records with source identifiers.

\subsection{Physics Problem Formalization}

% \ligeng{@xiongjing, please check the below contexts}
We formalize each physics problem as a Lean theorem before asking the worker to prove it. We use the open-source Science Mango pipeline \citep{science_mango}, which receives the problem text and figures. The pipeline first creates a blueprint that separates the physical laws, given values, results from earlier parts, and the claim to prove.

An autoformalization agent grounds this blueprint in existing formal mathematics by querying LeanExplore \citep{asher2025leanexplore}, a search engine for Lean 4 declarations, over Mathlib \citep{mathlib2020} and PhysLean \citep{toobysmith2025heplean}. It then writes typed Lean declarations with sorry proof bodies. It preserves the roles and dimensions of physical quantities. When no library definition exists, the agent introduces a small local abstraction that retains the physical meaning.  The target equation is withheld before being presented to agents for solving. If the model cannot write the physical laws as equations, it first produces a natural-language derivation and then autoformalizes that derivation. A physics reviewer compares the Lean statement with the original problem and rejects missing laws, omitted figure parameters, tautologies, and statements that assume the answer. Once the statement compiles and passes review, it is frozen. The prover may fill the proof body but cannot weaken the statement; semantic errors return for redrafting.

\subsection{Verifier}
% \ligeng{what is diff btw evaulter and VERIFIER?}

The verifier determines whether the current candidate satisfies a machine-evaluable acceptance condition. Its exit status and diagnostic log are retained as runner evidence. When verification fails, the calling agent receives the diagnostic summary, and the candidate cannot be accepted.

For mathematical tasks and formalized physics task, the verifier is Comparator \citep{lean_comparator}.  %\ligeng{what is the verifier for other tasks?}
, which checks untrusted Lean submissions against trusted challenge statements. Comparator builds the candidate in a sandbox, confirms that the submitted theorem matches the trusted challenge, restricts permitted axioms, and validates the exported proof with the Lean kernel. Every mathematical problem is encoded as a Lean theorem and counts as solved when Comparator accepts the candidate proof. The reviewer remains responsible for detecting reward hacking and problem misunderstanding; the verifier checks whether the stated theorem is proven. For coding task, the worker and reviewer can create test cases and execute the code to see if the time limit is satisfied and if the answer is correct.   

The reviewer and verifier play different roles. The reviewer is an agent in the worker--reviewer loop: it assesses the artifact semantically, identifies reward hacking or problem misunderstanding, and decides whether another revision is needed. The verifier is an objective, task-specific tool rather than an agent in the loop. Both the worker and reviewer can call it when they need machine-evaluable evidence about the current candidate.

\section{Experiments}
\label{sec:experiments}

% Compact signed deltas used in comparison tables.  Keep the sign outside the
% argument so gains and losses are formatted consistently.
\newcommand{\gain}[1]{\tiny\textsubscript{\textcolor{green!60!black}{+#1}}}
\newcommand{\loss}[1]{\tiny\textsubscript{\textcolor{red!80!black}{$-$#1}}}
\newcommand{\nochange}[1]{\tiny\textsubscript{\textcolor{black!55}{#1}}}

\begin{table}[t]
  \centering
  \setlength{\tabcolsep}{3.5pt}
  \renewcommand{\arraystretch}{1.15}
  \small
  \caption{HOA results across three evaluation levels. HOA reaches the 2026 gold threshold on IMO, IPhO, IChO, and IOI; solves all 672 PutnamBench problems \citep{tsoukalas2024putnambenchs}; and improves Lean-Eval from GPT-5.6's official 112/304 to 251/304, ranking \#1 \citep{leaneval}. The Lean kernel verifies every mathematics result. \textcolor{green!50!black}{\ding{51}} marks an open-source release, and \textcolor{red}{\ding{55}} marks a closed-source system. %\ligeng{add open source model results} \ligeng{add goedel architect}
  }
  \label{tab:main}
  \begin{tabular}{llcr}
    \toprule
    Benchmark & System & Open Source & Solved / Total \\
    \midrule
    \rowcolor{black!10} \multicolumn{4}{c}{High-School-Level Competitions---Commercial Models GPT-5.6-Sol} \\
    IMO 2026 & HOA (ours) & \textcolor{green!50!black}{\ding{51}} & 6 / 6 \\
    IPhO 2026 & HOA (ours) & \textcolor{green!50!black}{\ding{51}} & 30 / 30 \\
    IChO 2026 & HOA (ours) & \textcolor{green!50!black}{\ding{51}} & 97.2 / 100 \\
    IOI 2026 & HOA (ours) & \textcolor{green!50!black}{\ding{51}} & 600 / 600 \\
    \midrule
    \rowcolor{black!10} \multicolumn{4}{c}{High-School-Level Competitions---Open Models Kimi-K3} \\
    IMO 2026 & HOA (ours) & \textcolor{green!50!black}{\ding{51}} &   6 / 6 \\
    IPhO 2026 & HOA (ours) & \textcolor{green!50!black}{\ding{51}} &  30 / 30 \\
    IChO 2026 & HOA (ours) & \textcolor{green!50!black}{\ding{51}} & 97.1 / 100  \\
    IOI 2026 & HOA (ours) & \textcolor{green!50!black}{\ding{51}} &  600 / 600 \\
    \midrule
    \rowcolor{black!10} \multicolumn{4}{c}{College Level Competitions} \\
    PutnamBench & HOA (ours) & \textcolor{green!50!black}{\ding{51}} & \textbf{672 / 672} \\
    PutnamBench & Harmonic \citep{harmonic2025aristotle} & \textcolor{red}{\ding{55}} & \textbf{672 / 672} \\
    PutnamBench & Aleph Prover \citep{logicalintelligence2026aleph} & \textcolor{red}{\ding{55}} & \underline{668 / 672} \\
    PutnamBench & Mistral Prover \citep{mistral2026leanstral} & \textcolor{green!50!black}{\ding{51}} & 658 / 672 \\
    PutnamBench & Goedel-Architect \citep{chung2026goedelarchitectstreamliningformaltheorem} & \textcolor{green!50!black}{\ding{51}} & 642 / 672 \\
    PutnamBench & Seed-Prover 1.5 \citep{seedprover15} & \textcolor{red}{\ding{55}} & 581 / 672 \\
    \midrule
    \rowcolor{black!10} \multicolumn{4}{c}{Research Level Problems} \\
    Lean-Eval & HOA (ours) & \textcolor{green!50!black}{\ding{51}} & \textbf{251 / 304} \\
    Lean-Eval & AxiomProver \citep{axiom2026prover} & \textcolor{red}{\ding{55}} & \underline{246 / 304} \\
    Lean-Eval & Tau (caj.al) \citep{cajal2026tau} & \textcolor{red}{\ding{55}} & 200 / 304 \\
    Lean-Eval & Aristotle (Harmonic) \citep{harmonic2025aristotle} & \textcolor{red}{\ding{55}} & 162 / 304 \\
    Lean-Eval & Seed-Prover \citep{seedprover2025} & \textcolor{red}{\ding{55}} & 156 / 304 \\
    Lean-Eval & GPT-5.6 official \citep{openai2026gpt56} & \textcolor{red}{\ding{55}} & 112 / 304 \\
    \bottomrule
  \end{tabular}
\end{table}

\vspace{-0.2em}
\subsection{Experimental Setup}

We evaluate HOA across tasks that produce different artifacts and use different verification signals.

\paragraph{Benchmarks.}
Our mathematics evaluation spans three levels. At the high-school olympiad level, we use Axiom Math's formalized versions of the six IMO 2026 problems \citep{axiommath2026imo} with a stricter binary scoring rule. A solution receives full credit if it is mathematically correct and formalized as a Lean proof accepted by Comparator and the Lean kernel; otherwise, it receives zero points. We compare the total with the gold-medal cutoff. At the undergraduate level, we use the frozen 672-problem Lean subset of PutnamBench \citep{tsoukalas2024putnambenchs}. At the research level, we use the 302 formal theorem-proving tasks in Lean-Eval. PutnamBench and Lean-Eval use the same Comparator and Lean-kernel acceptance rule.

We also evaluate physics, chemistry, and competitive programming on the official 2026 IPhO, IChO, and IOI problems. IPhO and IChO responses are graded by an LLM judge according to their official scoring rubrics, and human reviewers check the resulting grades. IOI submissions are graded by Codeforces against hidden test cases. We compare each competition score with its 2026 gold-medal cutoff. The ablation study additionally uses a 50-problem Physics Cup subset.

\paragraph{Models and baselines.}
Our main closed-source model is GPT-5.6. We compare a standalone model (pure api call), a standalone agent (with codex harness), a standalone agent with wiki and verifier, and full HOA. Cross-model experiments use the open-weight Hugging Face checkpoints \texttt{moonshotai/Kimi-K3}, \texttt{zai-org/GLM-5.3}, and \texttt{deepseek-ai/DeepSeek-V4-Pro} \citep{kimik3model,glm53model,deepseekv4promodel}. We also compare with public Seed-Prover and AxiomProver results when the task set and acceptance rule match.
\paragraph{Metrics.}
For binary tasks, we report solved count and pass rate. For scored competitions, we report official raw points and whether the score reaches that year's gold threshold. When the sample size permits, we report bootstrap confidence intervals over problems. Failures, timeouts, and budget exhaustion remain in the denominator.

\vspace{-0.2em}
\subsection{Main Results}

% GPT 5.6 HOA -> based on 5.6 in caption
Table~\ref{tab:main} shows consistent gains across all three evaluation levels. \textbf{HOA meets or exceeds the 2026 gold thresholds on IMO, IPhO, IChO, and IOI} and solves \textbf{672/672 PutnamBench problems}. On Lean-Eval, HOA solves \textbf{251/304 problems (82.6\%)}, compared with \textbf{112/304 (36.8\%)} for the official GPT-5.6 result---a gain of \textbf{139 proofs and 45.7 percentage points}. Every formal proof is checked by the Lean kernel. Although the scoring rules differ across fields, the same pattern holds: review and revision convert failed attempts into accepted solutions.

\subsection{An Open Model with an Open Harness Can Match Closed-Source Agents}

\begin{table*}[t]
  \centering
  \setlength{\tabcolsep}{3.5pt}
  \renewcommand{\arraystretch}{1.15}
  \small
  \caption{Cross-model results for HOA and pure Codex on the 2026 IOI, IMO, IPhO, and IChO benchmarks. HOA improves performance across base models, with Kimi K3 + HOA nearly matching GPT-5.6 + HOA. Colored subscripts show absolute changes from pure Codex.}
  \label{tab:openclosed}
  \begin{tabular}{lllll}
    \toprule
    Model & IOI & IMO & IPhO & IChO \\
    \midrule
    \rowcolor{black!10} GPT-5.6-sol {\scriptsize(size undisclosed)} & & & & \\
    \quad + /Goal & \underline{562/600} & \textbf{6/6} & \textbf{30/30} & 87.8 \\
    \quad + HOA & \textbf{600/600}\gain{38} & \textbf{6/6}\nochange{0} & \textbf{30/30}\nochange{0} & \textbf{97.2}\gain{9.4} \\
    \midrule
    \rowcolor{black!10} Kimi K3 {\scriptsize(2.8T)} & & & & \\
    \quad + /Goal & 400/600 & \underline{4/6} & \textbf{30/30} & 65.0 \\
    \quad + HOA & \textbf{600/600}\gain{200} & \textbf{6/6}\gain{2} & \textbf{30/30}\nochange{0} & \underline{97.1}\gain{32.1} \\
    \midrule
    \rowcolor{black!10} DeepSeek V4 Pro {\scriptsize(1.6T)} & & & & \\
    \quad + /Goal & 347/600 & \underline{4/6} & 28.4/30 & 58.7 \\
    \quad + HOA & 411/600\gain{64} & \underline{4/6}\nochange{0} & 28.8/30\gain{0.4} & 60.7\gain{2.0} \\
    \midrule
    \rowcolor{black!10} GLM-5.3 {\scriptsize(744B)} & & & & \\
    \quad + /Goal & 249/600 & 2/6 & 28.3/30 & 54.8 \\
    \quad + HOA & 428/600\gain{179} & \underline{4/6}\gain{2} & \underline{29.9/30}\gain{1.6} & 58.8\gain{4.8} \\
    \bottomrule
  \end{tabular}
  \vspace{-10pt}
% Delete + wiki + verifier 
% OPen model
% Lean eval is too hard, IMO has so few problems
\end{table*}

\begin{figure}[t]
  \centering
  % Crop the second (lower) chart, excluding the source table and embedded caption.
  % graphicx trim order: left, bottom, right, top; preserve the original PDF vectors.
  \includegraphics[width=\linewidth,trim=152bp 96bp 64bp 734bp,clip]{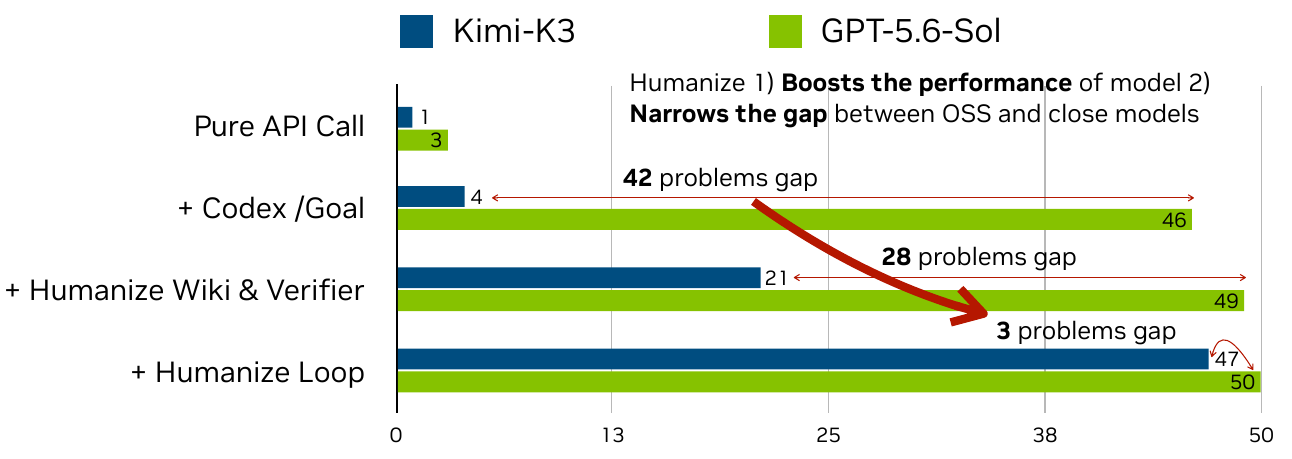}
  \caption{PutnamBench solved counts across four system levels for Kimi K3 and GPT-5.6 Sol on the first 50 problems. HOA raises Kimi K3 from 1/50 to 47/50 and GPT-5.6 Sol from 3/50 to 50/50. The gap between the models narrows from 42 problems with Codex \texttt{/Goal} to 3 problems with the full Humanize loop.
  }
  \label{fig:putnam-accuracy-progression}
\end{figure}

Table~\ref{tab:openclosed} shows that HOA consistently improves performance across base models. At the pure Codex level, the open models trail the closed GPT-5.6 model on most competitions; with HOA, Kimi K3 achieves full marks on IOI, IMO, and IPhO and nearly matches GPT-5.6 on IChO. Figure~\ref{fig:putnam-accuracy-progression} shows the same convergence on PutnamBench. Overall, the gap between open-model + HOA and closed-model + HOA is substantially narrower than the gap between their pure-Codex counterparts.
\subsection{Ablation Study}

We compare three levels of scaffolding---a direct API call, the original Codex harness, and the full HOA workflow---with GPT-5.6, Kimi K3, GLM-5.3, and DeepSeek V4 Pro. Table~\ref{tab:ablation} groups the results by base model and reports solved counts on PutnamBench and Physics Cup. HOA gives the best result for every model on both evaluations. The PutnamBench results for these configurations remain provisional.

\begin{table}[t]
  \centering
  \setlength{\tabcolsep}{3.5pt}
  \renewcommand{\arraystretch}{1.15}
  \small
  \caption{Model-, tool-, and flow-level results on 50-problem subsets of PutnamBench and Physics Cup. HOA performs best for every base model on both evaluations; colored subscripts show gains over the model-level API baseline.}
  \label{tab:ablation}
  \begin{tabular}{lrr}
    \toprule
    Model / Level & PutnamBench & Physics Cup \\
    \midrule
    \rowcolor{black!10} \multicolumn{3}{l}{GPT-5.6-sol} \\
    \quad {\scriptsize Model-Level (API)} & 3 & 31 \\
    \quad {\scriptsize Tool-Level (CLI)} & 46\gain{43} & 41\gain{10} \\
    % \quad {\scriptsize Tool-Level (CLI) + wiki + verifier} & 49\gain{46} & 41\gain{10} \\
    \quad {\scriptsize {Flow-Level (Humanize)}} & \textbf{50}\gain{47} & \textbf{44}\gain{13} \\
    \midrule
    \rowcolor{black!10} \multicolumn{3}{l}{Kimi-K3} \\
    \quad {\scriptsize Model-Level (API)} & 1 & 35 \\
    \quad {\scriptsize Tool-Level (CLI)} &  4\gain{3} & 37\gain{2} \\
    % \quad {\scriptsize Tool-Level (CLI) + wiki + verifier} &  21\gain{20} & 37\gain{2} \\
    \quad {\scriptsize Flow-Level (Humanize)} &  \textbf{47}\gain{46} & \textbf{42}\gain{7} \\
    \midrule
    \rowcolor{black!10} \multicolumn{3}{l}{GLM-5.3} \\
    \quad {\scriptsize Model-Level (API)} & 0 & 24 \\
    \quad {\scriptsize Tool-Level (CLI)} &  2\gain{2} & 34\gain{10} \\
    % \quad {\scriptsize Tool-Level (CLI) + wiki + verifier} &  21\gain{21} & 34\gain{10} \\
    \quad {\scriptsize Flow-Level (Humanize)} &  \textbf{25}\gain{25} & \textbf{40}\gain{16} \\
    \midrule
    \rowcolor{black!10} \multicolumn{3}{l}{DeepSeek V4 Pro} \\
    \quad {\scriptsize Model-Level (API)} & 0 & 32 \\
    \quad {\scriptsize Tool-Level (CLI)} &  4\gain{4} & 35\gain{3} \\
    % \quad {\scriptsize Tool-Level (CLI)+ wiki + verifier} &  8\gain{8} & 35\gain{3} \\
    \quad {\scriptsize Flow-Level (Humanize)} &  \textbf{13}\gain{13} & \textbf{39}\gain{7} \\
    \bottomrule
  \end{tabular}
  \vspace{-10pt}
\end{table}

\subsection{Analysis}

\begin{figure}[t]
  \centering
  \includegraphics[width=0.8\linewidth]{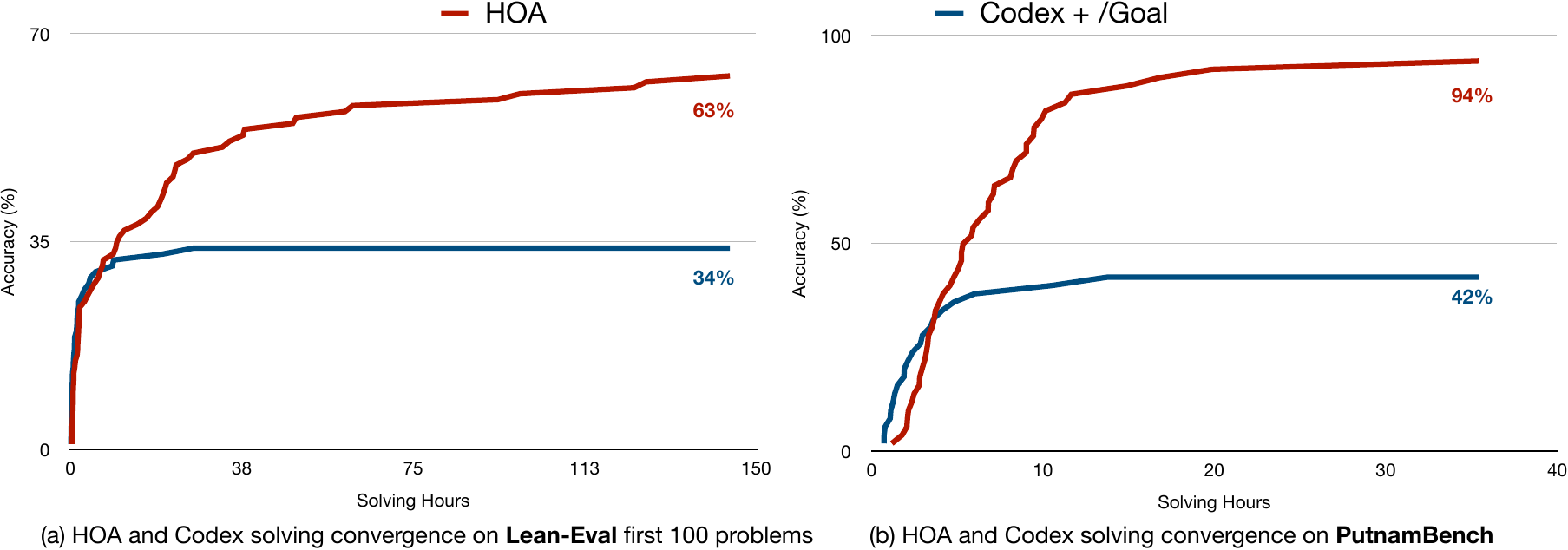}
  \vspace{-10pt}
  \caption{Cumulative accuracy versus worker time for GPT-5.6 on the 100-problem Lean-Eval subset (left) and Kimi K3 on the 50-problem PutnamBench subset (right). On Lean-Eval, Pure Codex + \texttt{/Goal} plateaus at 34/100 after about 30 hours, while HOA reaches 63/100 over 150 hours. HOA provides another scaling dimension: sustained agent time. Its review and revision loop converts longer runs into verified solutions by recovering from intermediate failures. All failures remain in the denominator. HOA's full Lean-Eval result is 251/304 (82.6\%), ranking \#1.
  %\ligeng{zhengyang plz send me the source data in csv on wechat., fig4 and 5}
  }
  \label{fig:humanize-curves}
\end{figure}

Figure~\ref{fig:humanize-curves} compares cumulative Lean-Eval accuracy against worker time for HOA and Pure Codex with \texttt{/Goal}. The evaluation uses the first 100 eligible problems in alphabetical order and retains every failure. Pure Codex + \texttt{/Goal} reaches 34/100, while HOA continues recovering verified solutions and reaches 63/100.
% For every issue
% flagged by the reviewer, we record what kind of error it found.  We then check
% how often the reviewer finds real errors, misses errors, or approves an
% incorrect answer.  We also count revisions that make no progress or make a
% previously correct answer worse.  
We launch HOA and Codex for 150 hours in lean eval experiment. Pure Codex with \texttt{/Goal} gets saturated and makes no progress after about 30 hours, while HOA and continuouly make progress throughout the process. The worker--reviewer loop turns this additional time into verified solutions instead of letting the worker try the same method and always make the same mistake. % \juihui{HOA gets 150h vs.\ ${\sim}$30h for /goal, seems unfair? what does "stalled attempts" mean, i thought /goal will continue indefinitely alreday\zhengyang{What experiment to rush?. We can have the accuracy higher than the /goal with infini compute or pass at a lot}}

% \begin{figure}[t]
%   \centering
%   \includegraphics[width=\linewidth]{figure/kimi_humanize_vs_original_codex_time_first50.pdf}
%   \caption{Cumulative Kimi K3 accuracy versus time on the first 50 PutnamBench problems. Under the Comparator-audited first-turn protocol, HOA reaches 94\%, while the original Codex scaffold reaches 6\%. The iterative workflow continues converting longer trajectories into verified solutions after the original scaffold plateaus.  %\zhengyang{Question: This result's baseline (red) is for Codex without verifier or wiki, but the previous openai image is for codex with verifier and wiki. Change?}
% }
%   \label{fig:kimi-putnam-time}
% \end{figure}

\paragraph{Wiki reuse in a Lean trajectory.}
The $G_2$ tensor-square decomposition run shows how the mathematics wiki acts as an executable interface rather than an unstructured hint. During the trajectory, the agent retrieved three records that had passed Comparator and independent review. They formalize a 14-dimensional simple $G_2$ quotient, a 64-dimensional irreducible highest-weight module, and the classification of fourteen irreducible tensor-square types. The agent applied these Lean declarations directly when assembling the final theorem instead of reproducing their proofs. The resulting artifact passed Comparator and both the Nanoda and default Lean kernels. This example shows how verified lemma records turn completed subproblems into typed, provenance-tracked building blocks for later formalization steps.

\section{Conclusion}

Reliable long-horizon problem solving requires more than a promising initial idea. An agent must detect errors, acquire missing knowledge, preserve the original specification, and determine when the task is complete. HOA addresses these needs with a mandatory worker--reviewer repair loop, targeted wiki retrieval, and task-specific verification. Comparator checks mathematical candidates against trusted Lean statements, while programs retain their native execution tests.

HOA reaches gold-medal-level performance on the 2026 IMO, IPhO, IChO, and IOI problems and receives full credit on PutnamBench. On research-level mathematics, it improves GPT-5.6 from 112/304 to 251/304 solved Lean-Eval tasks, ranking \#1. An open-weight model with HOA also outperforms a closed-source standalone agent. These results show that workflow design is a major source of capability rather than a thin layer around the base model.

Future work will explore parallel workflows and divide-and-conquer strategies that split complex problems into smaller parts, solve them independently, and combine their results.

\section*{Limitations}

Our evaluation covers a finite set of olympiad, programming, and formal-mathematics benchmarks, so it does not establish that HOA generalizes to arbitrary scientific research or to domains without reliable task-specific verification. Several reported results use closed model snapshots whose behavior, availability, and training data are not fully documented; possible benchmark contamination therefore cannot be ruled out. The comparison between workflows may also be affected by differences in inference time, token budgets, tool access, and stopping behavior. In particular, longer multi-session runs consume more computation than a single-agent baseline.

Comparator verifies agreement with trusted Lean statements, but this guarantee depends on the correctness and fidelity of those formalizations. It does not by itself establish that an informal problem was translated correctly. For chemistry, physics, and programming tasks, automated checks likewise cover only the properties encoded by the evaluator. Finally, the current experiments do not isolate every component interaction or fully characterize reviewer false positives and false acceptances. These constraints limit both causal attribution and the strength of claims about reliability beyond the evaluated settings.

\subsection*{AI Use Statement}

In this work, we used generative AI tools only for language polishing, grammar checking, and generating portions of the code. The research ideas, methodology, experimental design, and interpretation of results were developed by the authors. We reviewed all AI-assisted text to ensure that it accurately conveyed our intended meaning. LLM-generated code was verified and tested for correctness by authors. We take full responsibility for the final content of this work, including all text, claims, and artifacts produced with the assistance of generative AI.

\subsection*{Ethics Statement}

The experiments use public benchmark and competition problems and involve no human subjects or personal data. We will audit evaluation items for possible training contamination and report the resulting uncertainty. Benchmark performance should not be interpreted as evidence of broad scientific autonomy
\subsection*{Reproducibility Statement}
We provide the HOA code supplement containing the workflow implementation, native agent runtime, benchmark integration, the evaluation inventory, prompts, orchestration code, reviewer schema, wiki snapshot and source manifest, verifier specification, model and tool versions, budgets, per-run logs, and generated Lean proof artifacts.
Section~\ref{sec:experiments} lists the details of every benchmark and model checkpoint.
%\zhengyang{Whether to include the code for lean eval before sync with axiom?}
%\ligeng{no lean solution, only the scripts to reproduce.}

\bibliography{references}
\bibliographystyle{iclr2027_conference}

\appendix
\section{Appendix}
\label{sec:appendix}

\subsection{Evaluation Inventory}
\label{app:datasets}

The camera-ready version will list every evaluated problem with its benchmark version, competition year, public release date, maximum score, HOA score, verifier type, Lean artifact identifier for mathematical problems, and contamination-audit status. This inventory makes claims such as ``perfect score'' and ``gold-medal-level'' precise.

% \subsection{Prompts, Models, and Budgets}
\label{app:settings}

% For every run, we will release the worker and reviewer system prompts, wiki query template, reviewer output schema, Comparator interface, maximum review cycles, token and tool budgets, timeout, model snapshot, decoding parameters, and software versions. We will also report whether failures were retried and how we separated infrastructure errors from problem-solving failures.

\subsection{Reviewer Record Template}

Each reviewer record follows the schema below:
\begin{verbatim}
verdict: accept | revise
findings:
  - location: <artifact span>
    type: <error category>
    evidence: <supporting evidence>
    request: <minimal requested repair>
\end{verbatim}
An accepted artifact has no unresolved findings. When a worker disputes a finding, the evidence and final adjudication remain in the log.

\subsection{Additional Results}

The final appendix will report per-problem outcomes, confidence intervals, iteration curves, token and tool distributions, Comparator confusion analysis, wiki retrieval audits, and representative trajectories. To avoid selective reporting, the case studies will include successful repairs, false reviewer alarms, false acceptances, and runs that exhaust their budgets.

\end{document}